# Predicting Louisiana Public High School Dropout through Imbalanced Learning Techniques

Marmar Orooji
*Department of Computer Science*
*Rice University*
Houston, USA
marmar.orooji@rice.edu

Jianhua Chen
*Department of Computer Science*
*Louisisna State University*
Baton Rouge, USA
jianhua@csc.lsu.edu

*Abstract* — This study is motivated by the magnitude of the problem of Louisiana high school dropout and its negative impacts on individual and public well-being. Our goal is to predict students who are at risk of high school dropout, by examining Louisiana administrative dataset. Due to the imbalanced nature of the dataset, imbalanced learning techniques including resampling, case weighting, and cost-sensitive learning have been applied to enhance the prediction performance on the rare class. Performance metrics used in this study are F-measure, recall and precision of the rare class. We compare the performance of several machine learning algorithms such as neural networks, decision trees and bagging trees in combination with the imbalanced learning approaches using an administrative dataset of size of 366k+ from Louisiana Department of Education. Experiments show that application of imbalanced learning methods produces good results on recall but decreases precision, whereas base classifiers without regard of imbalanced data handling gives better precision but poor recall. Overall application of imbalanced learning techniques is beneficial, yet more studies are desired to improve precision.

*Keywords* — *Cost-Sensitive Learning, High School Dropout Prediction, Imbalanced Classification*

## I. INTRODUCTION

Louisiana has maintained one of the highest school dropout rates in the US for many years. The Public Affairs Research Council of Louisiana (PAR, October 2011) estimates that one in six of every public high school students in the state drops out of school. Dropping out of high school is associated with profound negative consequences lasting well into adulthood, including impairment in behavioral (e.g., higher rates of substance use), psychological (e.g., higher rates of depression), and social functioning (e.g., strained relationships with parents, teachers, and peers). In addition to individual-level economic costs (e.g., reduced earning potential), high dropout rates also have a negative societal impact, such as reduced tax revenue and increased crime rates. Thus, high school dropout remains a serious and unresolved problem, especially for at-risk students and those living in vulnerable communities. This study is motivated by the magnitude of the problem of Louisiana high school dropout and its negative impacts on individual and public wellbeing. In this study, our goal is to predict students who are at risk of high school dropout by using different classification techniques in order to inform intervention programs to prevent dropouts and all the destructive costs.

We use an administrative dataset from Louisiana Department of Education containing Louisiana public school students' records from 1999-2000 to 2011-2012 school years. In this dataset, dropout is a binary attribute, which serves as our class for prediction. Positive class contains students who have dropped out, while negative class consists of those who have not dropped out. Since we aim at identifying students at risk of dropout for prevention purposes, we try to build a classifier with high prediction performance on the positive class. In Section II, we introduce our dataset in depth and explain the pre-processing and feature extraction steps we took for training a classifier.

In our dataset, the positive class is the minority class, with only 4 percent of instances. This raises the imbalanced class problem in prediction. Detecting a rare class is a challenging task for standard classifiers because they have a bias towards classes with large number of instances and tend to only predict the majority class [1]. In addition, the features of the minority class may possibly be treated as noise and be ignored accordingly. On the contrary, noise might be detected as the minority class, since they both are rare patterns in the data [2]. Thus, standard classifier algorithms have poor performance on identifying the rare class. However, misclassifying rare events can incur heavy costs in different applications. For instance, a natural disaster might rarely happen, but failing to predict such catastrophe may result in huge and even irreparable harm to earth, people, and lives. Therefore, empowering classifiers to predict rare events is of crucial importance in different domain areas such as credit card fraud detection, rare disease detection, predicting criminality behavior, etc. Different approaches exist to handle imbalanced class problem and improve classifiers in better predicting the minority class of interest. In-depth reviews have been conducted on recent methods for learning from imbalanced data [3][4]. In Section III, we describe three approaches of sampling, case weighting, and cost-sensitive learning, which we use to enhance the prediction of dropouts. We employ various classification algorithms along with the three imbalance handling approaches to be able to find the best classifier with the highest prediction performance on dropouts. We introduce such classification algorithms in Section IV.

Besides the challenge of detecting the rare class, some evaluation metrics can be misleading in imbalanced classifications [5]. For example, a classification may result in high overall accuracy or precision even without identifying any of the rare instances. Hence, a careful choice and definition of evaluation metric is needed to correctly evaluate the performance of classifiers on predicting the minority class. In Section V, we address the limitation of accuracy and define the metrics, which are well suited for imbalanced class prediction.

## II. DATASET PREPARATION

Our original dataset is provided by Social Research and Evaluation Center at Louisiana State University, outsourced from Louisiana Department of Education, which contains enrollment and disciplinary data for students in Louisiana public schools in each school year from 1999-2000 to 2011-2012. This dataset has underlain various data science research in social work domain [6][7][8]. The original data is split into two datasets of school enrollment data and school disciplinary data. Each dataset is in form of separate excel spread sheets, one for each school year. Enrollment data contains student's demographics, grade level, school and district information, number of days of enrollments or absences, dropout flag, homelessness flag, receiving free lunch flag, etc. Disciplinary data consists of student's disciplinary actions at school such as expulsion and suspension information (see Fig 1). We created a dataset appropriate for our study out of the original datasets, using SQL Server Management Studio. Fig 2 shows the steps we took to prepare our desired dataset for this study.

First, we integrated enrollment and disciplinary information of students by merging the two datasets in each school year based on students' social security numbers. The datasets contain students' social security number (SSN) as their unique identifiers along with students' name and birth date. However, such identifiers are not clean in the raw data. For instance, we found records with the same SSN but different name and birth date. Among such discrepancies, some were negligible and implied mistyping errors while others were significantly different in their names and birth dates. Therefore, we identified records with the same SSN but adequately different names and birth dates, and prevented such records to be merged as the records for the same student.

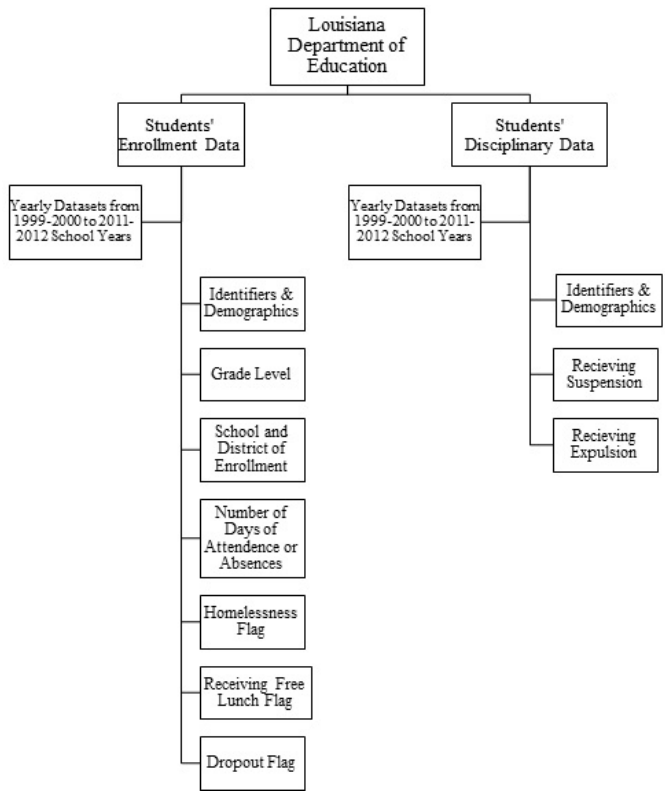

Fig 1. Original Datasets.

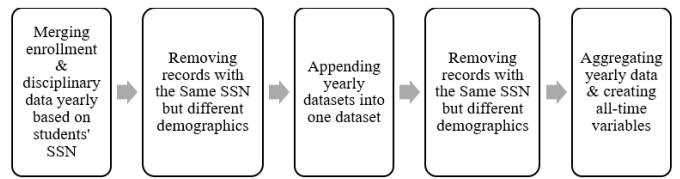

Fig 2. Steps to prepare our dataset.

After obtaining both enrollment and disciplinary data for students at each school year, we aimed at creating an all-time dataset, in which there should be one record for each student with the summarized information of all the school years. Therefore, we merged the datasets of all the 13 school years into one integrated dataset based on students' SSN and then aggregated their enrollment and disciplinary data to create all-time variables. In this step, again we prevented merging records with the same SSN but significantly different names and birth dates as the records for the same student. We found that we have records for 366,806 number of students in our dataset. We wanted to summarize students' academic performance and behavior during their enrollment in public schools and examine whether they dropped out of school in their last school year of enrollment. Thus, we created 18 variables, with their description shown in Table 1.

*Last Dropout Flag* is our target class, which is highly skewed. Only 4 percent of students (n=14,895) have dropped out while the remaining 96 percent (n=351,911) have not dropped out of high school.

We split our dataset into training set with 70 percent of instances (n=256,765) and test set with the remaining 30 percent (n=110,041). In both sets the distribution of *Last Dropout Flag* values remains the same.

## III. IMBALANCED LEARNING APPROACHES

### A. Resampling Techniques

Resampling techniques are used to balance the class distributions. Such techniques are applied on the training set, before building the learning model, in order to provide a balanced input data for classifiers. This resolves the challenge of learning from a minority class for standard classifiers. Therefore, classifiers can be modeled on a resampled balanced data to be able to learn about instances in the majority class on the original data. However, a careful consideration needs to be taken to obtain an honest estimate of the model performance when using resampling techniques. Although the classifier is modeled on a resampled training data, its prediction performance needs to be evaluated on the original imbalanced data to truly indicate its performance on detecting the minority class. Therefore, both validation and test set must remain imbalanced for evaluation.

For use of k-fold cross-validation in imbalance classification, large k is not recommended, since it produces higher imbalance folds. Therefore, in our study, we use 5-fold cross-validation for training the classifiers. For each of the five iterations of validation, we apply resampling on the four folds of training and keep the validation fold imbalanced, as shown in Fig 3.

TABLE 1. DESCRIPTION OF VARIABLES IN OUR ALL-TIME DATASET

| Attribute | Description | Type |
|---|---|---|
| Last Grade | Last grade level enrolled | Ordinal<br>-1 = pre-k;<br>0 = kindergarten;<br>1-12 =<br>$1^{st}$ to $12^{th}$ school grade. |
| Last Age | Age at the last enrollment | Numeric |
| Sex Cd | Gender | Binary<br>F = female;<br>M = male |
| Ethnic Cd | Ethnicity | Ordinal<br>1 = American Indian or Alaskan Native;<br>2 = Asian or Pacific Islander;<br>3 = Black;<br>4 = Hispanic;<br>5 = White;<br>6 = mixed race. |
| Fail Flag | Did the student ever fail one or more grades? | Binary<br>"Y" or "N" |
| Move Ahead Flag | Was the student ever moved ahead a grade or more? | Binary<br>"Y" or "N" |
| On Track Flag | Was the student always on track to graduate? | Binary<br>"Y" or "N" |
| Failed More than 2 | Did the student ever fail 2 or more grades? | Binary<br>"Y" or "N" |
| Avg Aggr Days Enrl Cnt | The average number of school days the student was enrolled in a school year | Numeric |
| Avg Aggr Days Abs Cnt | The average number of school days the student was absent in a school year | Numeric |
| Avg School Changes | The average number of times per year that a student changes school during a school year | Numeric |
| Avg District Changes | The average number of times per year that a student changes school district during a school year | Numeric |
| Ever Homeless | Was the student ever homeless? | Binary<br>"Y" or "N" |
| Ever Truancy Flag | Did the student ever receive 5 or more unexcused absences in any one month? | Binary<br>"Y" or "N" |
| Ever Free Lunch | Did the student ever receive fully subsidized (free) lunch? | Binary<br>"Y" or "N" |
| Ever Suspension | Did the student ever receive a suspension of any kind? | Binary<br>"Y" or "N" |
| Ever Expulsion | Did the student ever receive an expulsion of any kind? | Binary<br>"Y" or "N" |
| Last Dropout Flag | Did the student drop out in their last school year of enrollment? | Binary<br>"Y" or "N" |

One advantage of resampling techniques is that they improve imbalanced data prediction without a need to modify the classifier's algorithm. Thus, these techniques are adaptable to the use of any kind of classifier. In this study, we employ three resampling techniques: Random Down Sampling, Random Up Sampling, and Hybrid Methods. Random down sampling method randomly subsets the majority class to reduce the size to be equal to the size of minority class. Random up sampling technique randomly up samples (sampling with replacement) the minority class such that the size will be increased to match the size of majority class. The hybrid method we use in this study is Synthetic Minority Over-sampling Technique (SMOTE) [9] integrated with random down sampling. This technique down samples the majority class and generates new synthetic cases by interpolating two existing cases from the minority class, instead of duplicating minority samples.

*B. Case Weighting Approach*

In this approach, we assign normalized weights to instances. Classifiers, based on their algorithm, take such case weights into consideration when they are being trained. For instance, Decision Trees integrate case weights into their measure of impurity, while Support Vector Machines count such weights in the information loss term of their objective function. Generally, considering case weights in the learning algorithm makes the classifier put more emphasis on learning about instances with higher weights and thereby better predicting such instances. This approach is specifically applicable in imbalanced data classification to better learn from the rare class by assigning higher case weights to instances in the minority class. Instances with higher weights can be viewed as being duplicates, which is similar to employing up sampling.

In this study, we assign weights to each instance based on the class it belongs. The weights are inversely proportional to the frequency of the class, so that the instances in the minority class would have higher weights.

*C. Cost-Sensitive Learning*

Standard classifiers aim at minimizing the error rate, which is the rate of incorrect prediction of class labels. They do not make difference between incorrect predictions on different classes. In other words, they assume that misclassifying instances of any class costs the same. However, in many real-world applications, especially imbalanced data, correctly classifying instances of a class might be of crucial importance compared to other classes. Therefore, misclassifying instances of such class incurs higher cost than other misclassifications. For example, in medical diagnosis of cancer, classifying a patient with cancer in a "non-cancer" group, may result in death, while classifying a healthy person in a "cancer" group, may only result in conducting some medical tests and finally finding that the patient is not diagnosed with cancer.

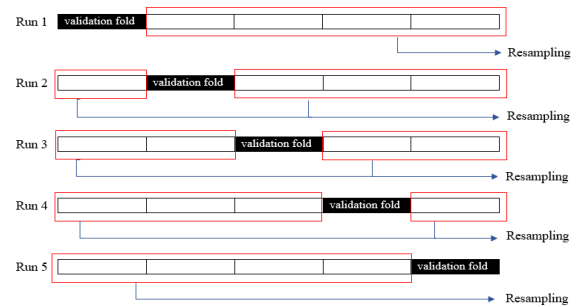

Fig 3. 5-fold cross-validation with resampling

Cost-sensitive learning is a classification method, which allows considering different costs for misclassifying instances into different classes. This technique aims at minimizing the overall misclassification cost. Consider a binary classification, Fig. 4 shows the confusion matrix and cost matrix. Confusion matrix contains the count of instances in each category. For instance, $f_{FP}$ is the number of negative instances classified as positive. Cost matrix shows the misclassification cost associated with each category. For example, $C_{FN}$ is the cost of misclassifying a positive class into a negative class. Although the two categories of TP and TN contain correctly classified instances and there should be no associated misclassification cost, sometimes negative numbers are assigned to these categories to reduce the overall cost. Having the confusion and cost matrix, the overall cost is calculated as:

$$Cost = (f_{TP} \times C_{TP}) + (f_{FN} \times C_{FN}) + (f_{FP} \times C_{FP}) + (f_{TN} \times C_{TN}) \quad (1)$$

In imbalanced-data, often misclassifying the rare class as majority is more expensive than misclassifying majority as rare class. As described before, standard classifiers do not consider misclassifications differently and also they perform poorly in predicting the rare class. Hence, integrating cost-sensitive learning approach in standard classifiers' algorithm enables them to better predict the rare class. Assuming the rare class as the positive class, $C_{FN}$ is assigned a large number so that a high cost incurs when it gets misclassified. In our experiments, we tested three cost values of 10, 100, and 1000 for $C_{FN}$.

In this study, we employ the integration of cost-sensitive learning into Decision Tree. Cost-sensitive Decision Tree uses the misclassification costs directly in tree construction, by minimizing the cost reduction instead of minimizing the impurity measure in attribute selection.

## IV. CLASSIFICATION ALGORITHMS

In this study, we use three standard classifiers, which are Single Decision Tree, Bagging Trees, and Artificial Neural Network. As describe earlier, such classifiers with their original algorithms cannot learn from the minority class. Therefore, in our experiments, we employ imbalanced learning approaches with such classifiers to enhance their prediction on the rare class. In this section, we briefly describe the three classification algorithms we use with their corresponding R packages.

### A. Single Decision Tree

In our study, we train a single decision tree model through the CART algorithm. "rpart" package in R has implemented such algorithm, and we use this package in our experiment. CART algorithm has two main steps:

- Step 1: Grow a full tree

  Recursively binary splitting the training data with the best split in each iteration which incurs the minimum impurity (Gini index used in "rpart"), until meeting the minimum leaf size or having no improvement in reducing the impurity.

- Step 2: Apply cost complexity pruning on the fully grown tree

  There is a complexity parameter ($cp$), which tunes the misclassification error versus the tree size. Based on the given $cp$, the tree is pruned to a subtree which minimizes $misclassification\ error + cp\ \times tree\ size$

In our experiment, we tune the complexity parameter to find the best pruned tree.

### B. Bagging Trees

Bagging is an ensemble method that combines models to reduce the variance and thereby enhance the prediction. It fits a classification model to $B$ number of bootstrap samplings (sampling with replacement) from the training data, and combines the prediction of the $B$ number of models by taking the majority vote class for each test data point. Bootstrap sampling is an approximation of what would happen if the population is resampled.

Bagging is recommended to use with unstable models, which are the models that greatly change with the small changes on the input data. Unpruned trees are example of unstable models. Therefore, in our study, we use bagging with unpruned trees, through "C5.0" package in R. C5.0 is the extension of C4.5 classification algorithm. In our experiment, we tune the number of bootstrap samplings ("trials" parameter in "C5.0" package) to get the highest performance model.

### C. Artificial Neural Network

Artificial Neural Network is a model of interconnected nodes with weighted links. The input layer consists of input nodes each corresponds to a variable. The output layer contains output nodes representing the classes. Hidden layers comprising artificial neurons can appear in between to transfer input data to output.

In this study, we use "nnet" package in R, which implements a feed-forward neural network with a single hidden layer. Each neuron in the hidden layer is a function of linear combination of inputs and the weighted links to the neuron. Accordingly, each output is modeled as a linear combination of neurons and the weighted links to the output. The activation function used in this package is the sigmoid.

In our experiment, we tune two parameters in our implemented neural network algorithm. One is the number of neurons in the hidden layer ("size" parameter in "nnet" package) and the other is weight decay ("decay" parameter in "nnet" package) to avoid overfitting and regularizing the model.

|  | | PREDICTED CLASS | |
|---|---|---|---|
| | Confusion Matrix | Yes | No |
| ACTUAL CLASS | Yes | $f_{TP}$ | $f_{FN}$ |
| | No | $f_{FP}$ | $f_{TN}$ |

|  | | PREDICTED CLASS | |
|---|---|---|---|
| | Cost Matrix | Yes | No |
| ACTUAL CLASS | Yes | $C_{TP}$ | $C_{FN}$ |
| | No | $C_{FP}$ | $C_{TN}$ |

Fig 4. Confusion Matrix vs. Cost Matrix

## V. EVALUATION METRICS

Confusion matrix is used to define various performance measures to evaluate the classifiers. Having *Last Dropout Flag* as our class, the confusion matrix is shown in Fig. 4. One of the most widely used evaluation metric is Accuracy. However, accuracy is misleading in imbalanced-data classification. For instance, in our dataset, 14,895 out of 366,806 number of students have dropped out. Even if none of the 14,895 dropped out students has correctly classified (i.e., all the students classified as not being dropped out), accuracy would be 95.94 percent. Therefore, in this study, we use Precision, Recall, , and F-measure as our evaluation metrics, which are defined as:

$$Precision = \frac{f_{TP}}{f_{TP} + f_{FP}} \quad (2)$$

$$Recall = \frac{f_{TP}}{f_{TP} + f_{FN}} \quad (3)$$

$$F - measure = \frac{(1 + \beta^2) \times precision \times recall}{(\beta^2 \times precision) + recall} \quad (4)$$

Precision shows what percentage of the predictions on the positive class is correct. Recall indicates the proportion of the positive class instances that are correctly classified. High precision implies low $f_{FP}$ while high recall shows low $f_{FN}$. As described before, in this study (similar to other rare class classifications), classifying a positive class as negative is more expansive than the opposite way. In other words, we want to minimize the misclassification of a dropped out student, which is the FN error. Therefore, in our experiments, we favor models with high recall than high precision. F-measure is a combined evaluation metric integrating both precision and recall, with a parameter (β) to put weight on either precision or recall. Assigning β as one implies that the precision and recall have the same level of importance in showing the performance of the classifier. Choosing β less than one puts more weight on precision while β more than one treats recall as the more important metric in evaluation. In our experiments, we set β as five, to emphasize on recall metric. We call such measure F5 score. Precision, recall, and F5 score in all of our experiments in this study are calculated with respect to the positive class, which is the rare class (Dropout = Yes).

## VI. EXPERIMENTS

The experiments are conducted on a machine with Intel(R) Core(TM) i7-2600 CPU (3.40 GHz) and 16 GB RAM. SQL Server Express 2014 has been used to create and store our dataset. Classification algorithms and imbalanced learning approaches are all programmed with R 3.5.1 software.

We employed the three classification algorithms, described in Section IV, through R packages on our training data to build the models. To improve the performance of the classifiers, we use sampling, case weighting, and cost-sensitive learning approaches along with the classifiers. Totally we have built 17 models, i.e., 6 models of Single Decision Tree, 6 models of Bagging Trees (standard, with down sampling, with up sampling, with SMOTE, with case weighting, with cost-sensitive learning), and 5 models of Artificial Neural Network (standard, with down sampling, with up sampling, with SMOTE, with case weighting).

### A. Model Parameters Tuning

Each classifier has a set of parameters that needs to be tuned to result the best model in terms of the highest performance measure. We tuned the parameters for all the 17 models to select parameter values, which result the best model in terms of the highest cross-validation F5 score. For parameter tuning, we used "caret" package in R.

In training a single decision tree, we tried four complexity parameter values of 0, 0.0001, 0.0005, and 0.001 on our models. We tuned the number of bootstrap sampling in bagging trees. In C5.0 package of R, this parameter is named trials. We tried four different trials: 1, 5, 10, and 50. There are two parameters in our single hidden layer feed-forward neural network model, which require tuning. One is the number of hidden units and the other is the weight decay, which is the regularization parameter to avoid overfitting. Larger number of hidden units results in more complex model and thereby increase the risk of overfitting. We tested the number of hidden units to be 1, 3, and 5, and the weight decay to be 0 (no regularization), 0.0001, and 0.1. In employing cost-sensitive learning approach, we tried three cost values of 10, 100, and 1000 for $C_{FN}$. The results of parameter tuning are shown in Table 2.

TABLE 2. MODELS PARAMETER TUNING RESULTS

| Model | Tuned Parameters | Train CV F5 |
|---|---|---|
| Single Decision Tree | cp = 0 | 0.234 |
| Single Decision Tree – Down Sampling | cp = 0.0005 | 0.76 |
| Single Decision Tree – Up Sampling | cp = 0.0005 | 0.757 |
| Single Decision Tree – Hybrid Sampling | cp = 0.001 | 0.721 |
| Single Decision Tree – Case Weighting | cp = 0.0001 | 0.748 |
| Single Decision Tree – Cost Sensitive Learning | cp = 0.0001 $C_{FN}$ = 1000 | 0.747 |
| Bagging Trees | trials = 5 | 0.252 |
| Bagging Trees – Down Sampling | trials = 50 | 0.773 |
| Bagging Trees – Up Sampling | trials = 1 | 0.437 |
| Bagging Trees – Hybrid Sampling | trials = 1 | 0.695 |
| Bagging Trees – Case Weighting | trials = 1 | 0.754 |
| Bagging Trees – Cost Sensitive Learning | trials =1 $C_{FN}$ = 1000 | 0.744 |
| Neural Network | size = 3 decay = 0.1 | 0.072 |
| Neural Network – Down Sampling | size = 5 decay = 0.1 | 0.763 |
| Neural Network – Up Sampling | size = 3 decay = 0.0001 | 0.752 |
| Neural Network – Hybrid Sampling | size = 3 decay = 0.1 | 0.760 |
| Neural Network – Case Weighting | size = 3 decay = 0.0001 | 0.700 |

## B. Results on the Test Set

After tuning parameters of our classifiers by selecting models with highest 5-fold cross-validation F5 scores, we built our classifiers with the tuned parameters on the training data. Then, we employed such models on the test set and evaluated their performance.

To compare the prediction performance of our models, Fig 5 illustrates precision and recall measures of our models on the test set. The models are sorted by recall values in descending order. It is indicated that cost sensitive learning approach and case weighting technique outperform the sampling techniques in terms of returning the highest recall values on all models. We can see that the three standard classifiers that are built on the imbalanced data are the three models with the lowest recall values. This indicates the fact that standard classifiers are not well performed in detecting the rare class. Among the three standard classifiers built on the imbalanced data, bagging trees model has the highest recall value. In our experiment, this model builds five decision trees on five bootstrap samplings, and it shows that such sampling with replacement reduces the skewness of the data. This implies the potential effectiveness of ensemble methods on detecting minority class.

Another finding about standard classifiers is that their precision decreases when they are employed with the imbalanced learning approaches. This denotes that imbalanced learning approaches aim at minimizing the incorrect classification of the rare class into the majority class (decreasing FN), at the cost of increasing the incorrect classification of the majority class into the minority class (increasing FP). As described before, this result is acceptable in many real-world rare classification problems (same as our dropout classification) since incorrectly classifying the minority class is greatly expensive.

## VII. Conclusions

In this study, we aimed at predicting students who are at risk of high school dropout. We examined a large-scale student record dataset provided by Louisiana Department of Education. Only 4 percent of our dataset contained students who have dropped out of school. This created the problem of imbalanced classification, i.e., the standard classifiers cannot detect the rare class. To address this problem, we studied the imbalanced learning techniques, such as sampling, case weighting, and cost-sensitive learning, to enhance the prediction performance on our target class.

The results have shown that the recall values of our models have been increased substantially when imbalanced learning techniques are employed while the precision values have been decreased. This implies that reducing FN error increases FP error. However, FN error is much more expensive than FP error. Therefore, we aimed at reducing FN error, which results in increasing the recall values. The highest recall value is obtained by the cost-sensitive single decision tree (91.6%), followed by the neural network trained on the case weighted data (91%) and cost-sensitive bagging decision trees (90.9%). Among the standard classifiers trained on the imbalanced data, bagging trees model outperforms in detecting the rare class, which indicates the effectiveness of ensemble methods in imbalance classification.

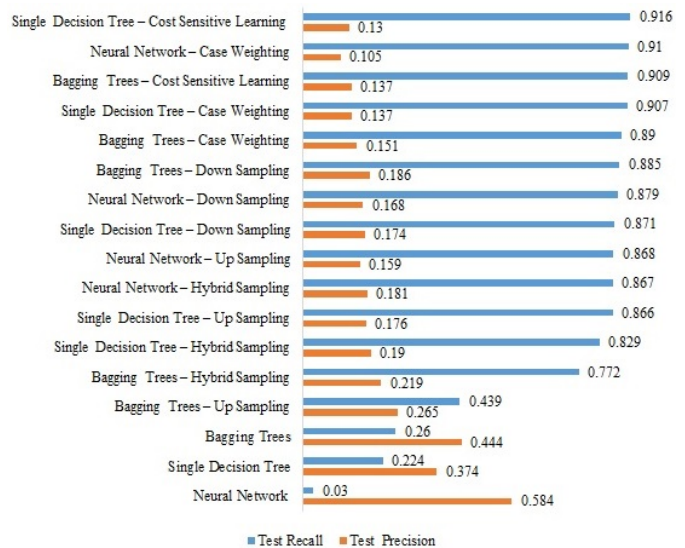

Fig 5. Comparing precision and recall of models on test set

### Acknowledgment

The authors would like to acknowledge the Social Research and Evaluation Center at Louisiana State University for providing an administrative Louisiana Educational dataset.